\documentclass[final]{cvpr}

\usepackage{times}
\usepackage{epsfig}
\usepackage{graphicx}
\usepackage{amsmath}
\usepackage{amssymb}
\usepackage{caption}
\usepackage{subcaption}
\usepackage{multirow}
\usepackage{array}
\usepackage{array, makecell}
\usepackage{booktabs}

\usepackage[pagebackref=true,breaklinks=true,colorlinks,bookmarks=false]{hyperref}

\def\cvprPaperID{****} 

\def\cvprPaperID{****} 
\def\confYear{CVPR 2021}

\begin{document}

\title{To Perceive or Not to Perceive: Lightweight Stacked Hourglass  Network}  

\author{Jameel Hassan Abdul Samadh, Salwa K. Al Khatib\\
{\tt\small\{jameel.hassan, salwa.khatib\}@mbzuai.ac.ae}}

\maketitle
\thispagestyle{empty}


\section{Introduction}
Human pose estimation (HPE) is a classical task in computer vision that focuses on representing the orientation of a person by identifying the positions of their joints. HPE can be used to understand and analyze geometric and motion-related information of humans. The stacked-hourglass architecture presented by Newell et al. in \cite{hourglass} is one of the first compelling deep learning-based approaches to HPE, as classical approaches dominated HPE literature prior to it. In this work, repeated bottom-up and top-down processing is utilized to capture information from various scales and intermediate supervision is introduced to iteratively refine predictions at each stage. This led to a significant boost in accuracy as compared to state-of-the-art approaches at the time. \par 

However, HPE is meant to be a real-time application as it is often used as a precursor to another module. Thus, focus on computational efficiency is vital in this context. In this study, we implement architectural and non-architectural modifications on the stacked hourglass network to obtain a model that is both accurate and computationally efficient.

In what follows, we provide a brief description of the baseline model. The original architecture is made up of multiple stacked hourglass units, each of which is composed of four downsampling and upsampling levels. At each level, downsampling is achieved through a residual block and a max pooling operation, while upsampling is achieved with a residual block and naive nearest neighbor interpolation. This process ensures that the model captures both local and global information, which is important to coherently understand the full body for an accurate final pose estimate. After each max pooling operation, the network branches off to apply more convolutions through another residual block at the pre-pooling resolution, the result of which is added as a skip connection to the corresponding upsampled feature map in the second half of the hourglass. The output of the model is a heatmap for each joint that models the probability of a joint's presence at each pixel. Intermediate heatmaps after each hourglass are predicted upon which a loss is applied. In addition, these predictions are projected to a larger number of channels and act as the input to the subsequent hourglass, along with the input of the current hourglass and its feature map outputs. The source code and trained models are available at \href{https://github.com/jameelhassan/PoseEstimation}{https://github.com/jameelhassan/PoseEstimation}

\section{Design Choices}

\subsection{Depthwise Separable Convolutions}
\begin{figure}[ht]
    \centering
    \includegraphics[width=0.55\linewidth]{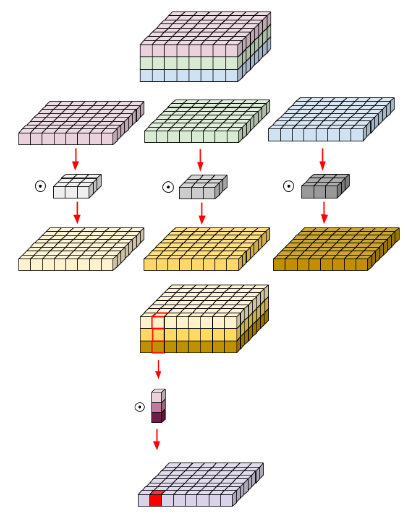}
    \caption{\centering Depthwise separable convolution}
    \label{fig:separable}
\end{figure}

Depthwise separable convolutions replace traditional convolutions to reduce the number of parameters of the convolution operation. This is performed by splitting the convolution using convolution spatially across the channels individually and then aggregating channel information through pointwise convolutions as in Figure \ref{fig:separable}.

\subsection{Dilated Convolution}
A dilated convolution, described in Eq.~\ref{eq:dilated}, is a variant of a regular convolution operation that has the capacity to exponentially increase the receptive field  without losing resolution or coverage, as is the case with pooling operations. 

\begin{equation}
    \label{eq:dilated}
    (F*_{l}k)(p) = \sum_{s+lt=p}F(s)k(t)
\end{equation}

where $k$ is a discrete filter, $l$ is the dilation factor, and $*_{l}$ is an $l$-dilated convolution operation. A regular convolution corresponds to a 1-dilated convolution. Dilated convolutions have little to no effect on computational complexity \cite{mask3d}.

\subsection{Ghost Bottleneck}
\begin{figure}[bth]
    \centering
    \includegraphics[width=0.7\linewidth]{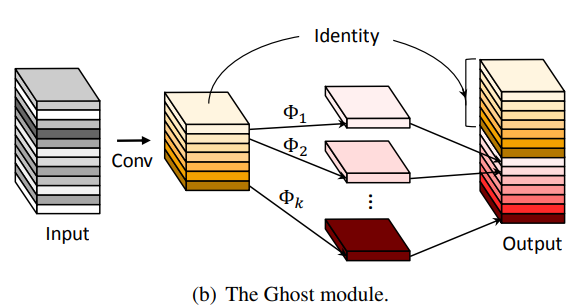}
    \caption{Ghost bottleneck}
    \label{fig:ghost}
\end{figure}

The Ghost bottleneck proposed by \cite{ghostnet} also reduces the computational complexity of convolution operations by splitting the convolution differently. In order to produce a fixed number of channels, the Ghost bottleneck outputs a fraction of the channels using regular convolutions and the rest are produced through cheaper linear operations as in Figure~\ref{fig:ghost}. These are concatenated and convolved to output the required number of channels.  

\subsection{DiCE Bottleneck}
A Dimension-wise
Convolutions for Efficient Networks (DiCE) unit is a convolutional unit proposed by Mehta et al. in \cite{dicenet} that compromises dimension-wise convolutions followed by dimension-wise fusion. The convolution operation is applied across each of the three input dimensions (width, height, and depth). To combine the encoded information along each of these dimensions, an efficient fusion unit is used to combine these representations. Thus, a DiCE unit can efficiently capture information along the spatial and channel dimensions. 

\begin{figure}
    \centering
    \includegraphics[width=\linewidth]{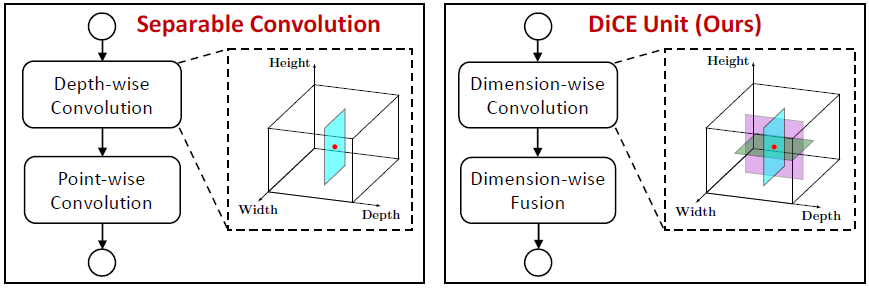}
    \caption{\centering The DiCE unit introduced in \cite{dicenet} compared to separable convolutions}
    \label{fig:dicenet}
\end{figure}

\subsection{Shuffle Bottleneck}
The shuffle unit, first presented in \cite{shufflenet},  uses pointwise \textbf{group} convolutions and channel shuffling to both boost computational efficiency and maintain accuracy. 

\subsection{Perceptual Loss}
The perceptual loss is used to compare similar images with minor differences. Here, we use it as a feature level mean-squared error (MSE) loss between the two images, which computes the loss at a high-level feature map instead of the original image space. The assumption made here is that if the first hourglass is made to 'perceive' what the second hourglass 'perceives' at that high feature level, the overall performance of the network will improve. The total loss, presented in Equation~\ref{eq:loss}, consists of the perceptual loss and the original prediction losses with higher weight for the prediction losses.
\begin{equation}
    \label{eq:loss}
\mathcal{L} = \lambda \big (\alpha(\mathcal{L}_{HG1} + \mathcal{L}_{HG2})+ (1-\alpha)(\mathcal{L}_{percep}) \big )
\end{equation}

\subsection{Residual connections}
We also replace the existing addition of residual connections with concatenated residual connections followed by a pointwise convolution to obtain the required number of channels, referred to as $ResConcat$. We also include a residual connection from the narrowest feature map of the hourglass (the neck) to the next hourglass neck, referred to as $NarrowRes$.

\begin{table*}[bht]
\label{tab:results}
\caption{\centering Summary of results. The reported validation PCKh scores are computed on the 2,958 images in the validation set of MPII dataset}

        \resizebox{1\linewidth}{!}{\begin{tabular}{l c c c c c c c c c c c c} 
        & \multicolumn{8}{c}{\textbf{Validation PCKh (\%)}} & \multicolumn{3}{c}{\textbf{Compute}} & \multirow{2}{*}{\textbf{Weighted Mean (\%)}} \\ 
        \cmidrule(l){2-9} \cmidrule(l){10-12}
        \textbf{Method} & Head & Shoulder & Elbow & Wrist & Hip & Knee & Ankle & Mean & \shortstack{Training time} & MAdds & Params & \\ 
        \midrule 
            \multicolumn{12}{c}{\textbf{\rule{0pt}{2pt}Baseline}}\\\midrule
        1 HG & 57.44 & 67.85 & 65.57 & 60.44 & 68.89 & 45.90 & 24.56 & 56.95 & - & 5.90G & 3.58M & 4.46\\ %
        \textbf{2 HG} & 58.12 & 69.67 &69.08 & 63.35 & 74.88 & 49.20 & 26.38 & 59.76 & 2h 31m & 9.14G & 6.7M & - \\
        3 HG & 57.91 & 70.30 & 69.81 & 64.59 & 75.33 & 50.21 & 28.81 & 60.61 & - & 12.38G & 9.87M & -7.07\\
        \midrule
            \multicolumn{12}{c}{\textbf{\rule{0pt}{2pt}{Alternative Bottlenecks}}} \\
        \midrule
        Ghost bottleneck & 56.68 & 65.62 & 61.96 & 53.45 & 67.44 & 41.93 & 23.17 & 54.03 & 2h 32m &4.03G & 2.53M & 4.16\\ 
        Shuffle bottleneck &  55.66 & 64.76 & 61.96  &  55.16&  65.64& 41.95& 22.08&53.65  & 2h 39m& 4.10G &  \textbf{0.94M} & 5.94\\ 
        DiCE bottleneck & 56.65 & 68.27 & 66.85&  61.16 & 72.39 &48.66& 25.91& 58.26 &- & - &3.2M & -\\ 
        \midrule
            \multicolumn{12}{c}{\textbf{\rule{0pt}{2pt}{Dilated Convolutions}}}\\
        \midrule
        Dilated bottleneck &  57.57   &    69.29    &69.17&    64.16 & 75.42   &50.59&    27.19   & \textbf{60.20} &  2h 44m & 15.29G & 16M & -3.35\\ 
        Dilated bottleneck + separable & 57.88       & 69.34   & 68.11 &   62.50 & 73.93 &  49.36&    26.41 &  59.35 &  2h 40m  & 6.78G &2.9M & 7.72\\ 
        Multidilated in HG & 57.54 &  69.43 & 68.69 &   63.99 & 74.12 &50.29 & 27.35 & 59.89 & 2h 53m &7.63G & 4.2M & 5.59\\ 
        Multidilated everywhere &  57.88&  69.29 & 68.72 &   63.90  &73.78  & 48.92  &26.83 &59.57 & 2h 50m   & 5.76G &3.8M & 7.8\\ 
        \midrule
            \multicolumn{12}{c}{\textbf{\rule{0pt}{2pt}{Modified Baseline}}} \\\midrule
        Weighted $L_{percept}$ & 56.62 & 66.58 & 62.79 & 57.99 & 66.54 & 42.65 & 20.81 & 54.56 & - & 9.14G & 6.7M & -7.05\\
        Low channels (168-84) & 57.74 & 68.65 & 67.67 & 62.56 & 72.20 & 48.66 & 26.86 & 58.86 & - & 4.42G & 2.98M & 9.53\\ 
        Fully separable (168-84) & 57.61 & 69.31 & 68.54 & 62.91 & 73.98 & 49.53 & 25.77 & 59.36 & 2h 34m &\textbf{2.34G}&1.36M & 11.48\\ 
        \shortstack[l]{Separable $ResConcat$ (168-84) + \\$NarrowRes$ }&  \textbf{58.36} & \textbf{69.40} & \textbf{68.45} & \textbf{63.64} & \textbf{73.52} & \textbf{49.10} & \textbf{27.19} & \textbf{59.59} & \textbf{2h 35m} & \textbf{2.57G} & \textbf{1.41M} & \textbf{14.87}\\
        Separable ${ResConcat}$ (168-84) + \\$NarrowRes$ + $\mathcal{L}_{percept}$ & 56.72 & 66.00 & 63.39 & 57.78 & 67.25 & 43.76 & 23.00 & 55.10 & 2h 34m &\textbf{2.57G}&1.41M & 8.85\\
        \bottomrule 
        \end{tabular}}

        \label{tab:results} 
        \end{table*}

\section{Experiments and Results}
Here, we consider our baseline to be the 2-stack variant of the stacked hourglass architecture, which has a validation mean PCKh score of $59.76\%$. All the subsequent experiments were carried out with a learning rate of $10^{-3}$, RMSprop optimizer, and a batch size of 24 (unless otherwise mentioned) for 20 epochs on a Quadro RTX 6000 GPU (24 workers). The training and evaluation of the model are carried out using the MPII dataset \cite{mpii} as is done in \cite{hourglass}. This dataset includes a diverse set of ~25k labeled (joints, scale, and center) images of ~40k people. The images are resized to $256x256$ after having been cropped to exclusively include the person who is the target of the labeling. 

The evaluation metric used to capture the performance of a model is the percentage of correct key points (PCKh@0.5) which measures if the predicted joint and the true joint are within 50\% of the head-bone link. We bold the best model in the last row, and each of the best performance metrics out of all models.

\subsection{Alternative bottlenecks}
\textbf{Ghost Bottleneck:} The bottleneck in the original model is replaced using a ghost module instead of the $3\times3$ convolution layer in the bottleneck as shown in Figure \ref{fig:ghost}. This enables us to reduce the number of parameters by more than half, however the accuracy of the model also suffers. \par 

\textbf{Shuffle Bottleneck:} Here we replace the original bottleneck with the ShuffleNet bottleneck. This vastly drops the number of parameters to less than 1M but has similar performance and MAdd operations count to Ghost bottleneck as seen in Table \ref{tab:results}. \par 

\textbf{DiCE Bottleneck:} The replacement of the bottleneck with the DiCE bottleneck gives a considerable drop in the number of parameters to $3.2M$ with very minimal loss in the PCKh score. 
 
\subsection{Dilated convolutions}

\begin{figure}
    \centering
    \includegraphics[width=0.72\columnwidth]{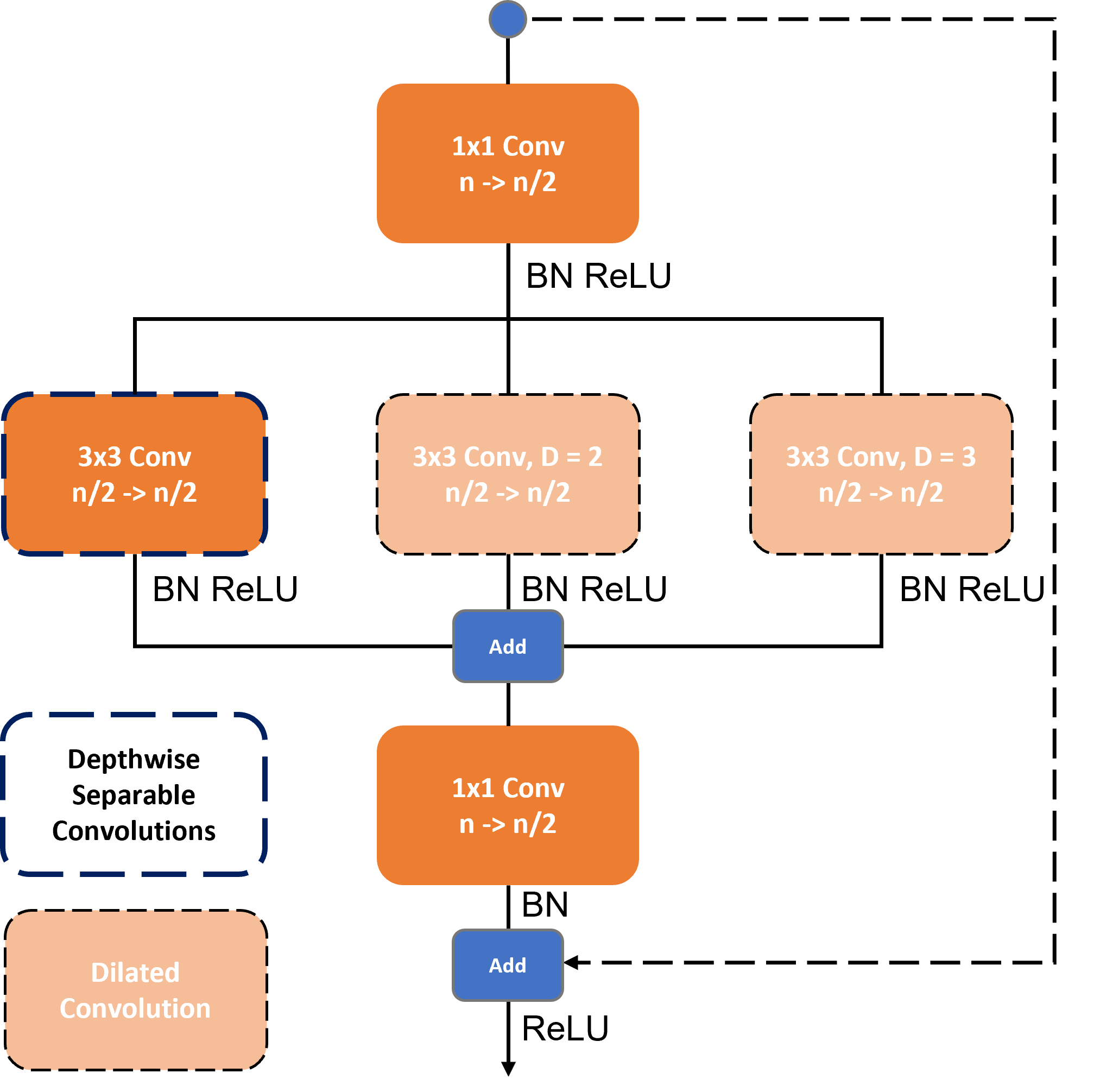}
    \caption{\centering Multi-dilated bottleneck}
    \label{fig:multidilated}
\end{figure}

Next, we experiment with dilated convolutions. We run an experiment with a bottleneck of two $3\times3$ dilated convolutions with a dilation factor of 2 and a residual connection. PCKh increases to $60.2\%$, so we deduce that dilated convolutions have the capacity to boost performance. We proceed by running the same experiment but with separable convolutions, which assist in reducing compute from $15.29G$ to $6.78G$ while also reducing accuracy to $59.35\%$. To further balance out accuracy with compute, we use a multi-dilated bottleneck as the one proposed in \cite{lightweight}, which uses 3 parallel $3\times3$ separable convolutions with a dilation factor of 1, 2, and 3 respectively, sandwiched in between two 1x1 convolutions. First, we use this bottleneck exclusively in the hourglasses while maintaining the original bottleneck elsewhere, which results in a model that is both \textbf{lighter} than the baseline ($4.2M$ parameters and $7.63G$ Multiplication and Addition Operations (MAdds)) and with \textbf{better performance} $(59.89\%)$. In the final experiment, we use the multi-dilated bottleneck as a replacement for all bottlenecks in the network, which causes a drop in accuracy and compute. It is worthy of note that we keep the number of channels throughout the different bottlenecks constant in the aforementioned experiments.

\begin{figure*}[ht]
\centering
\includegraphics[width=0.75\textwidth]{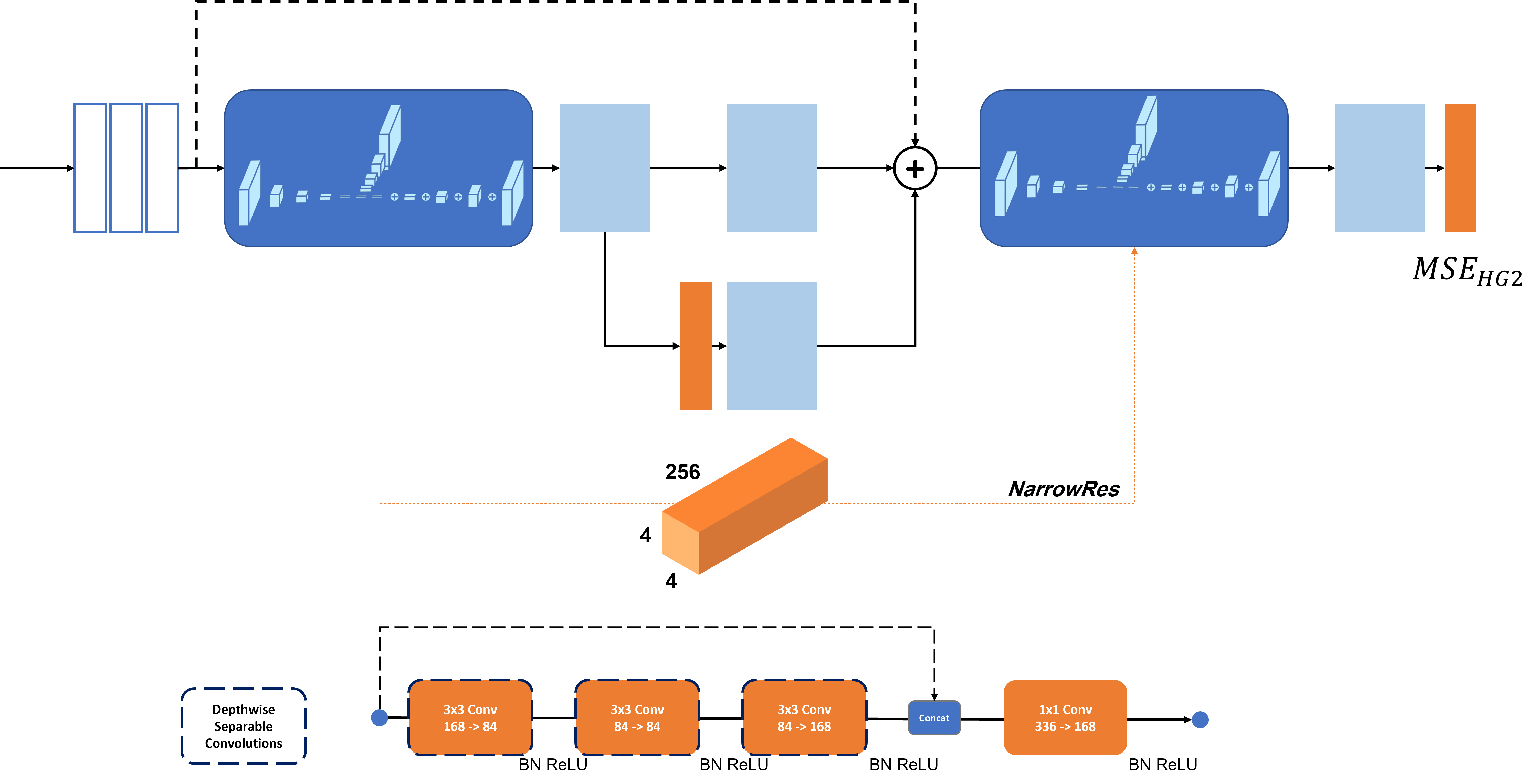}
    \label{fig:architecture}
    \caption{Architecture of the best model}
\end{figure*} 

\subsection{Modified baseline}
We then analyze the impact of different architectural and non-architectural changes on the baseline model. Specifically, we incorporate the following in a stagewise manner:
\begin{itemize}
    \item $\mathcal{L}_{percept}$
    \item Reduced number of channels
    \item Separable convolutions
    \item $ResConcat$
    \item $NarrowRes$
\end{itemize}

We initially incorporate the perceptual loss to the baseline. However, this drops the accuracy quite considerably by a margin of $5.2$. Since we want to achieve a lighter model, we then reduce the number of channels in the bottleneck layers. We find an optimum setting as $168-84$ (from $256-128$) from the baseline. This drops the accuracy marginally while having less than half the MAdds and number of parameters. We then incorporate fully separable convolutions for the bottleneck which further drops the number of parameters and MAdds while also giving an improvement of $0.5$ in the PCKh score. \par

We then incorporate the $ResConcat$ and $NarrowRes$ residual connection changes to the model. This gives an improvement of $0.23$ with a minor increase in the number of params and MAdds. This addition accounts for our best model so far. We then include the perceptual loss which decreases the PCKh score by $4.49$. Here, we scaled the weighed perceptual loss (weight $\alpha=0.3$) by a factor $\lambda=2$. We hypothesize that since the loss is now smaller with weights of $0.7$ and $0.3$, we will require a higher learning rate or a scaling factor on the loss. Perhaps tuning these parameters further might enable the perceptual loss to increase the accuracy.

\section{Best Architecture}

After analyzing the results of the different experiments, the choice of the best architecture still remains debatable. To determine the best architecture, we compute a metric that is a weighted average of three other metrics (last column of Table~\ref{tab:results}) that jointly reflect the tradeoff between accuracy and compute: \% change in the number of parameters, \% change in MAdd operations, and \% change in mean PCKh with respect to the baseline. Using this metric, we conclude that the final row architecture resulted in the best model that strikes a balance between accuracy and time towards the goal of a lightweight model.

Thus, the best architecture consists of two stacked hourglasses, with lower number of channels $(168-84)$ using depthwise separable convolutions. They also include $ResConcat$ and $NarrowRes$ skip connections with the scaled, weighted perceptual loss.

\section{Conclusion}
We design a lighter stacked hourglass network with minimal loss in performance of the model. The lightweight 2-stacked hourglass has a reduced number of channels with depthwise separable convolutions, residual connections with concatenation, and residual connections between the necks of the hourglasses. The final model has a marginal drop in performance with 79\% reduction in the number of parameters and a similar drop in MAdds. Adding the proposed perceptual loss did not help increase performance as what was originally thought; we hypothesize that using a better loss such as Kullback–Leibler divergence loss, which is not as harsh of a heuristic as MSE is, would make this approach perform well, i.e.,``perceive" better. 

{\small
\bibliographystyle{ieee_fullname}
\bibliography{egbib}
}


\end{document}